%% file: main.tex
\documentclass{article}

\usepackage{microtype}
\usepackage{graphicx}
\usepackage{booktabs} % for professional tables
\usepackage{xspace}
\usepackage{amssymb} % For checkmark and cross symbols
\usepackage{graphicx} % For \resizebox command
\usepackage{amsmath}
\usepackage{enumitem}
\setitemize{noitemsep,topsep=2pt,parsep=2pt,partopsep=0pt}
\usepackage{subcaption}
\usepackage{amsthm}
\usepackage{thm-restate}
\usepackage{multirow}
\newtheorem{thm}{\textbf{Takeaway}}
\usepackage[most]{tcolorbox}
\definecolor{light-gray}{gray}{0.95}
\tcolorboxenvironment{thm}{
   center,
   boxsep=0pt,
   top=0pt,
   bottom=0pt,
    width=\linewidth,
    colframe=light-gray,
    colback=light-gray
}

\usepackage{amssymb}
\usepackage{mathtools}

\DeclareMathOperator*{\argmin}{arg\,min}

\usepackage{hyperref}
\usepackage{textcomp}

\newif\ifcomments
\commentstrue 
\ifcomments
    \newcommand{\todo}[1]{{\color{red}{[TODO: #1]}}}
    \newcommand{\RK}[1]{{\color{blue}{\bf\sf [Redwan: #1]}}}
    \newcommand{\ankur}[1]{{\color{purple}{\bf\sf [Ankur: #1]}}}
    \newcommand{\sh}[1]{{\color{cyan}{\bf\sf [??: #1]}}}
\else
    \newcommand{\todo}[1]{}
    \newcommand{\RK}[1]{}
    \newcommand{\ankur}[1]{}
    \newcommand{\sh}[1]{}
\fi

\newcommand{\sys}{{\scshape FairServe}\xspace}

\usepackage[accepted]{mlsys2024}

\mlsystitlerunning{Ensuring Fair LLM Serving Amid Diverse Applications}
\begin{document}

\twocolumn[
\mlsystitle{Ensuring Fair LLM Serving Amid Diverse Applications}

% unless you've provided the [accepted] option to the mlsys2024


\mlsyssetsymbol{equal}{*}

\begin{mlsysauthorlist}
\mlsysauthor{Redwan Ibne Seraj Khan}{equal,vt,ms}
\mlsysauthor{Kunal Jain}{equal,ms}
\mlsysauthor{Haiying Shen}{ms,uva}
\mlsysauthor{Ankur Mallick}{ms}
\mlsysauthor{Anjaly Parayil}{ms}
\mlsysauthor{Anoop Kulkarni}{ms}
\mlsysauthor{Steve Kofsky}{ms}
\mlsysauthor{Pankhuri Choudhary}{ms}
\mlsysauthor{Ren\`ee St. Amant}{ms}
\mlsysauthor{Rujia Wang}{ms}
\mlsysauthor{Yue Cheng}{uva}
\mlsysauthor{Ali R. Butt}{vt}
\mlsysauthor{Victor R$\ddot u$hle}{ms}
\mlsysauthor{Chetan Bansal}{ms}
\mlsysauthor{Saravan Rajmohan}{ms}
\end{mlsysauthorlist}

\mlsysaffiliation{vt}{Department of Computer Science, Virginia Tech, Virginia, USA}
\mlsysaffiliation{uva}{Department of Computer Science, University of Virginia, Virginia, USA}
\mlsysaffiliation{ms}{Microsoft, Redmond, USA}

\mlsyscorrespondingauthor{Redwan Ibne Seraj Khan}{redwan@vt.edu}

\mlsyskeywords{Machine Learning, MLSys}

\vskip 0.3in

\begin{abstract}
In a multi-tenant large language model (LLM) serving platform hosting diverse applications, some users may submit an excessive number of requests, causing the service to become unavailable to other users and creating unfairness. Existing fairness approaches do not account for variations in token lengths across applications and multiple LLM calls, making them unsuitable for such platforms.
To address the fairness challenge, this paper analyzes millions of requests from thousands of users on MS CoPilot, a real-world multi-tenant LLM platform hosted by Microsoft.
Our analysis confirms the inadequacy of existing methods and guides the development of \sys, a system that ensures fair LLM access across diverse applications. 
\sys proposes %to incorporate 
application-characteristic aware request throttling %method 
coupled with a weighted service counter based scheduling technique to curb abusive behavior and ensure fairness. 
Our experimental results on real-world traces demonstrate \sys's superior performance compared to the state-of-the-art method in ensuring fairness. We are actively working on deploying our system in production, expecting to benefit millions of customers world-wide. 
\end{abstract}
]



\printAffiliationsAndNotice{\mlsysEqualContribution} % otherwise use the standard text.

\input{1_Introduction}
\input{2_Background}
\input{3_DataAnalysis}
\input{4_FairServe}
\input{5_Evaluation}
\input{6_RelatedWork}
\input{7_Conclusion}
%\section*{Acknowledgements}


\bibliography{mlsys_references}
\bibliographystyle{mlsys2024}

%%%%%%%%%%%%%%%%%%%%%%%%%%%%%%%%%%%%%%%%%%%%%%%%%%%%%%%%%%%%%%%%%%%%%%%%%%%%%%%
%%%%%%%%%%%%%%%%%%%%%%%%%%%%%%%%%%%%%%%%%%%%%%%%%%%%%%%%%%%%%%%%%%%%%%%%%%%%%%%
% SUPPLEMENTAL CONTENT AS APPENDIX AFTER REFERENCES
%%%%%%%%%%%%%%%%%%%%%%%%%%%%%%%%%%%%%%%%%%%%%%%%%%%%%%%%%%%%%%%%%%%%%%%%%%%%%%%
%%%%%%%%%%%%%%%%%%%%%%%%%%%%%%%%%%%%%%%%%%%%%%%%%%%%%%%%%%%%%%%%%%%%%%%%%%%%%%%
% \appendix
% \section{Please add supplemental material as appendix here}
% %
% Put anything that you might normally include after the references as an appendix here, {\it not in a separate supplementary file}. Upload your final camera-ready as a single pdf, including all appendices.

%%%%%%%%%%%%%%%%%%%%%%%%%%%%%%%%%%%%%%%%%%%%%%%%%%%%%%%%%%%%%%%%%%%%%%%%%%%%%%%
%%%%%%%%%%%%%%%%%%%%%%%%%%%%%%%%%%%%%%%%%%%%%%%%%%%%%%%%%%%%%%%%%%%%%%%%%%%%%%%

\end{document}

%% file: 1_Introduction.tex
\section{Introduction}
In recent years, multi-tenant personalized large language model (LLM) serving platforms, which support a wide range of applications like interactive question answering (QA), summarization, coding assistance, etc.,  have surged in popularity, with notable examples including Copilot~\cite{Copilot}, Punica~\cite{Punica}, and S-LoRA~\cite{sheng2024slora}. 
On these serving platforms, some users may flood the system with excessive requests, causing service disruptions for others and leading to considerable unfairness. 
For instance, on February 13, 2024, OpenAI reported a partial outage lasting over five hours for its API and ChatGPT services due to DDoS attacks, which recurred with varying durations in the following days ~\cite{LLMOutage}. Therefore, to maintain service availability and fairness, it is crucial to prevent abusive usage and ensure fair distribution of resources among users on LLM platforms.

Existing LLM platforms enforce limits on the number of requests each user can submit within a given period, such as requests per minute (RPM) to address the challenge of abuse prevention and ensuring fairness. 
For example, Google Gemini for Workspace limits usage to 500 times per month ~\cite{GoogleLimit}, and OAI Chat GPT Plus imposes limits of 80 messages every 3 hours on GPT-4o and up to 40 messages every 3 hours on GPT-4~\cite{OpenAILimit}.
However, such rate-limiting solutions have drawbacks. First, modern LLM applications are built from multiple LLM agents, each operating with an LLM model, working together to produce a single response for the user. The graph of interconnected LLM call requests is termed an LLM interaction. Throttling at the LLM request-level without %a clear 
understanding %of
the application's needs can result in incomplete user responses and resource wastage. Second, requests can still be throttled even if the system is underloaded, causing resource underutilization and user frustration.

The Virtual Token Counter (VTC) scheduler prioritizes requests of users who receive the minimal service when processing batches, aiming to prevent unfairness~\cite{vtc}. However, unlike RPM, VTC does not throttle or block users, thus making it unable to fully prevent abusive behaviors. Based on the principle of fair-queueing~\cite{nagle1987packet}, VTC gives both benign and abusive users equal service, leading to resource wastage, longer request queues, increased latency, and user frustration. 

\begin{table*}[ht]
    \centering
    \vspace{-0.1in}
    \caption{Comparison of \sys and current methods.
    }
    \resizebox{\textwidth}{!}{%
    \begin{tabular}{|l|c|c|c|c|c|c|c|}
        \hline
        Method &Maintains user & Avoid resource & Block abusive & User \& App characteristic-  & Reduce resource  & Consider multi-\\ 
        &  experience  & under-utilizations& behaviors& aware service calculation & wastage & agent LLM apps \\ 
        \hline
        RPM~\cite{GoogleLimit,OpenAILimit} & $\times$ & $\times$ & $\checkmark$ & $\times$ & $\times$ & $\times$\\ 
        \hline
        VTC~\cite{vtc}&  $\times$ & $\checkmark$ & $\times$ & $\times$ & $\times$ & $\times$\\ 
        \hline
        \sys & $\checkmark$ & $\checkmark$ & $\checkmark$ & $\checkmark$ & $\checkmark$ & $\checkmark$  \\ 
        \hline
    \end{tabular}%
    }
    \vspace{-0.15in}
    \label{tab:compareMethods}
\end{table*}

Additionally, modern LLM platforms receive requests from users having varying needs as they belong to different applications. Different applications have distinct typical ranges for input and output token lengths. For instance, article summarization applications often involve long inputs and short outputs, while code generation applications typically feature short inputs and long outputs. Hence, when scheduling requests for users across diverse apps, an ideal policy would be application-characteristic aware to ensure fairness and at the same time curb abusive behavior.

To gain a deep understanding of these problems, we conduct a comprehensive analysis on MS CoPilot, a leading real-world multi-tenant personalized LLM serving platform. Our investigation ($\S$~\ref{sec:data_analysis}) confirmed the limitations of existing methods and highlighted the need for a more sophisticated approach to ensure fairness amid diverse
applications with different properties. Based on these findings, we designed a novel system, \sys, which is tailored to ensure fair access to LLM resources amid diverse applications. 
\sys consists of two core components: (1) Overload and Interaction-driven Throttling (OIT); and (2) Weighted Service Counter (WSC) scheduler which collaborate together to serve user requests.

Unlike conventional throttling solutions, OIT throttles requests only when the system is overloaded, maximizing resource utilization. Additionally, it implements throttling at the LLM interaction-level (a single LLM interaction may consist of multiple simpler LLM calls), leveraging application characteristics to prevent token wastage and curbs abusive behavior of user across apps by setting application-specific limits. 
The WSC component complements OIT by selecting requests from the users who have received the least service, defined by a weighted resource slice. This weight is calculated based on the token ratio to ensure fairness across users.
Note that, Unlike the equality-centric approach proposed by VTC, our method prioritizes serving users who have received the least service, building on an equity-centric approach~\cite{demers1989analysis} that addresses the diverse needs of users across different applications. Table~\ref{tab:compareMethods} compares the effectiveness of the existing methods against \sys to prevent abusive behaviors and achieve fairness.

Specifically, this paper makes the following contributions.

\begin{itemize}[noitemsep]
    \item We conduct large-scale analysis on millions of requests across 34 applications on a leading multi-tenant LLM platform, revealing intricate characteristics of modern LLM applications and guiding future research.
    \item We introduce a novel approach, Overload and Interaction-driven Throttling (OIT), for throttling LLM requests to curb abusive behavior.
    \item We design a novel mechanism, Weighted Service Counter (WSC) for scheduling LLM requests to ensure fairness amid diverse applications.
    \item We present the design and implementation of \sys---a new LLM serving system that integrates WSC and OIT for fairness and abuse prevention.
    \item We integrate \sys in a open-source LLM serving platform~\cite{sheng2024slora} using state-of-the-art iteration-level continuous batching~\cite{yu2022orca} and compare against a series of policies for LLM serving fairness. Our results on a testbed of over a thousand users show that \sys reduces waiting queue delays by $10.67-93\times$, decrease latency (time-to-first-token) by $1.03-1.06\times$, increase throughput by $1.03-1.75\times$ (with 0\% token wastage) across apps while curbing abusive behaviour and enabling better service to users by $99.45-100\%$. 
\end{itemize}

We are actively working on deploying our system in production and making the source code of our prototype available to the research community to foster further research and development in this critical area.

%% file: 2_Background.tex
\section{Background}
\label{sec:background}

\noindent\textbf{Transformer Architecture.} Modern LLMs are built on multiple transformer blocks~\cite{vaswani2017attention}. Each block consists of an attention operation which applies three weight matrices $W_K$, $W_Q$, and $W_V$ respectively to the encoded represention of the input tokens, X of an user, to calculate the key K, query Q, and value V respectively associated with the corresponding input prompt. 

\vspace{-3mm}
{\small
\begin{align}
\begin{split}
 Q = XW^Q, K = XW^K, V = XW^V\\
    \text{Attention}(Q, K, V)=\text{softmax}(\frac{QK^T}{\sqrt{d_k}})V
\label{eq:atten}
\end{split}
\end{align}
}
\vspace{-3mm}

Next, the K, Q, and V matrices are used to calculate the attention head following Eq.~\ref{eq:atten}, which is later linearly combined with other attention heads to get the multi-head attention (MHA), allowing the model to focus on different parts of the sequence in different representational spaces. After the MHA block, the output undergoes layer normalization and is passed through a position-wise feed-forward network (FFN). The FFN consists of two linear transformations with an activation function in between, allowing further refinement of the information processed by the MHA block. The final output retains the same dimensionality as the input.

\noindent\textbf{LLM Inference Procedure.} LLM inference procedure consists of two phases--a prefill phase and a decode phase. In the prefill phase, the LLM processes all input tokens from the user in parallel in each iteration and generates the first output token.  Parallel processing makes efficient use of GPU resources. During input token processing, the context from those tokens are computed by the model's attention layers and cached in the KV cache~\cite{pope2023efficiently}. 

In the decode phase, the LLM autoregressively generates one output token at a time. The decode phase requires access to the KV cache of all previously processed tokens in order to perform the attention operation.
As the model predicts the next token, it appends the newly generated token to the original prompt and updates the KV cache, making it to grow linearly with the number of tokens generated. 
Once the first token is generated, subsequent token generation only requires the last generated token and the accumulated KV cache as inputs.
The decode phase ends when a special end-of-sentence (EOS) token is generated or the request reaches its pre-defined maximum length. 
The decode phase is primarily memory-bound with low compute utilization, since the model processes one token at a time.

\begin{figure}[t!]
\centering
\includegraphics[width=0.48\textwidth]{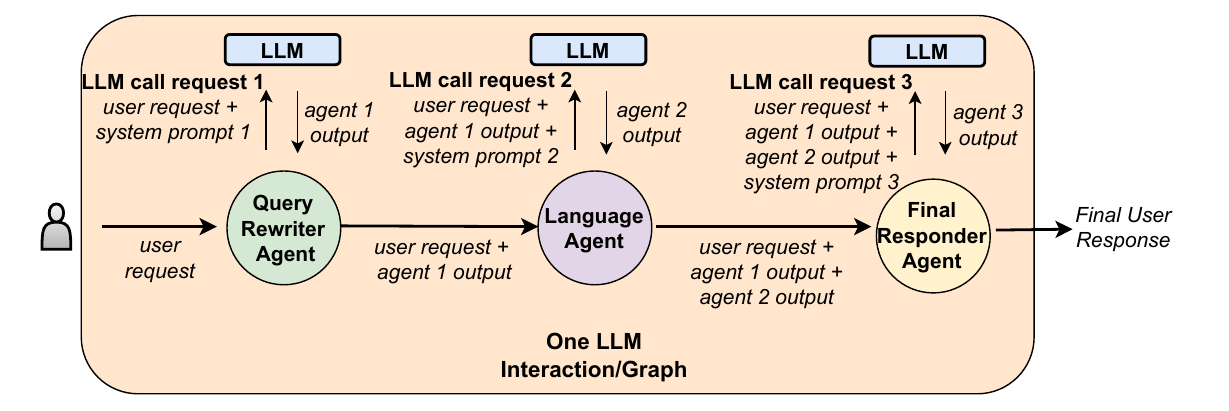}
\caption{Example of an LLM interaction. }
\label{fig:llmagents_illustration}
\end{figure}

\noindent\textbf{Multi-agent LLM inference.} Modern LLM inference applications comprises of complex LLM calls at multiple-levels to generate a response to a user query. User queries can be complex, requiring decomposition into simpler sub-queries. Each sub-query is addressed by an LLM agent utilizing a specialized LLM. Once the answers to these sub-queries are generated, they are aggregated, and a contextualized query is constructed. This refined query is subsequently processed by another LLM to produce the final response for the user. In short, multiple LLM agents, each in charge of a separate, simpler task, can collaborate together to perform a larger, but more complex task~\cite{Parrot2024,wu2023autogen,hong2023metagpt}. 
For example, Figure~\ref{fig:llmagents_illustration} shows how multiple LLM agents can collaborate together to generate a single user response. We denote the entire graph of LLM calls formed to generate a user response as one {\bf LLM interaction}.  {\it\bf System prompts} refer to pre-defined prompts or instructions sent to the language model to establish the context, rules, or behavior before processing a request. 

\noindent\textbf{LLM Inference in Multi-tenant Environments.} 
In multi-tenant environments, sequentially handling user requests increases latency and GPU underutilization, necessitating parallel processing through either (1) request-level batching or (2) iteration-level (continuous) batching.
In request-level batching, a batch is selected from waiting requests and processed until completion. Inference engines like FasterTransformer~\cite{fastertransformer} and Triton~\cite{triton} use this method, which simplifies operations but wastes GPU resources and increases wait times due to zero-padding shorter requests to match longer ones.

Request-level batching's resource wastage problem has been solved using iteration-level batching introduced by Orca~\cite{yu2022orca}. Iteration-level batching allows requests to dynamically enter and exit the current batch of requests in processing at the end of each model iteration. Modern LLM serving systems like vLLM~\cite{vllm}, LightLLM~\cite{lightllm}, and S-LoRA~\cite{sheng2024slora} use iteration-level batching to significantly increase system throughput and utilization. 

\noindent\textbf{Abuse Prevention in Multi-tenant Environments.}
High volumes of requests in modern LLM applications increase the risk of abusive behavior, potentially impacting fairness for other users. Rate-limiting solutions, such as RPM, can prevent abuse and ensure fairness by imposing limits on user request rates within a period of time~\cite{OpenAILimit}. However, in multi-agent scenarios, throttling a request in the middle of an interaction can lead to resource wastage. We discuss the implications of RPM solutions in $\S$~\ref{subsec:impact_rpmthrottling}.

A recent approach to mitigating unfairness
%abusive behavior 
is VTC~\cite{vtc}. VTC ensures fairness by allocating an equal number of tokens per minute (TPM) to all users, regardless of whether they are abusive or benign. This strategy treats every user equally by consistently serving the same TPM and refraining from throttling any user’s requests. However, we argue that this approach is inadequate for mitigating abusive behavior and ensuring fairness in complex multi-tenant, multi-agent environments. The diverse nature of applications leads to distinct user behaviors and resource demands, which require a more nuanced approach to ensure equitable service.  Furthermore, an abusive user might be a bot flooding the system with requests to degrade the service quality for benign users. Providing equal service to such users without throttling their requests consumes unnecessary resources and increases the response times for genuine users. 
We discuss implications of VTC approach in $\S$~\ref{subsec:impact_user} and $\S$~\ref{subsec:impact_application}. 

%% file: 3_DataAnalysis.tex
\section{Production Workload Analysis}\label{sec:data_analysis}

We analyze a full day of trace data from 34 distinct applications, covering millions of requests\footnote{We have anonymized the exact volume and application names due to confidentiality reasons.}.
We investigate the impact of user behavior, application behavior, LLM agents, and throttling techniques to gain insights into designing a fair LLM serving system. 
\subsection{Impact of User Behavior}
\label{subsec:impact_user}
\begin{figure*}[t!]
    \centering
    \begin{subfigure}{0.24\textwidth}
        \includegraphics[width=\textwidth]{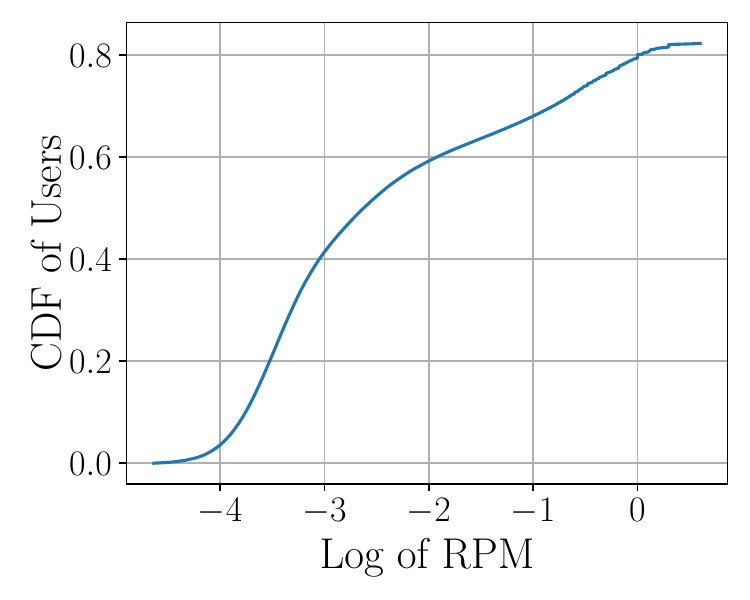}
        %\caption{Overall distribution of RPM across different users}
        \caption{Overall distribution of RPM across different users.}
        \label{fig:users_vs_rpm_new}
    \end{subfigure}
    \begin{subfigure}{0.24\textwidth}
        \includegraphics[width=\textwidth]{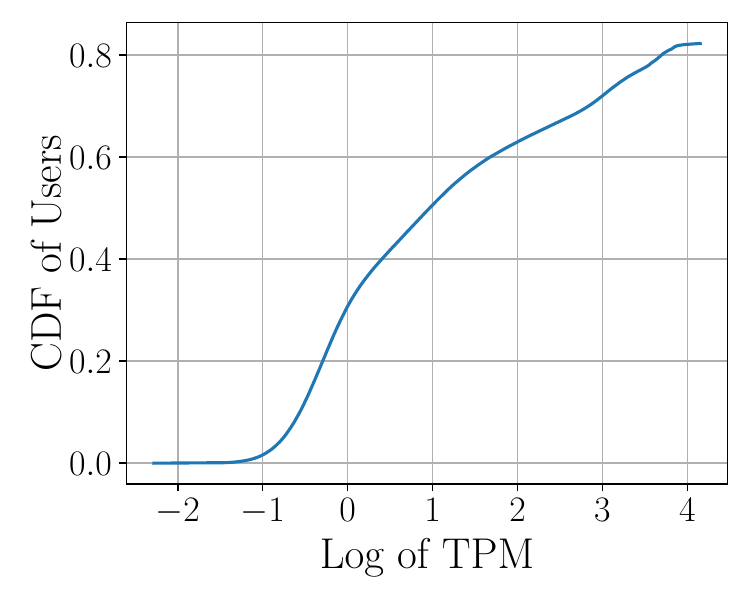}
        %\caption{Overall distribution of TPM across different users}
        \caption{Overall distribution of TPM across different users.}
        \label{fig:users_vs_tpm_new}
    \end{subfigure}
    \begin{subfigure}{0.24\textwidth}
        \includegraphics[width=\textwidth]{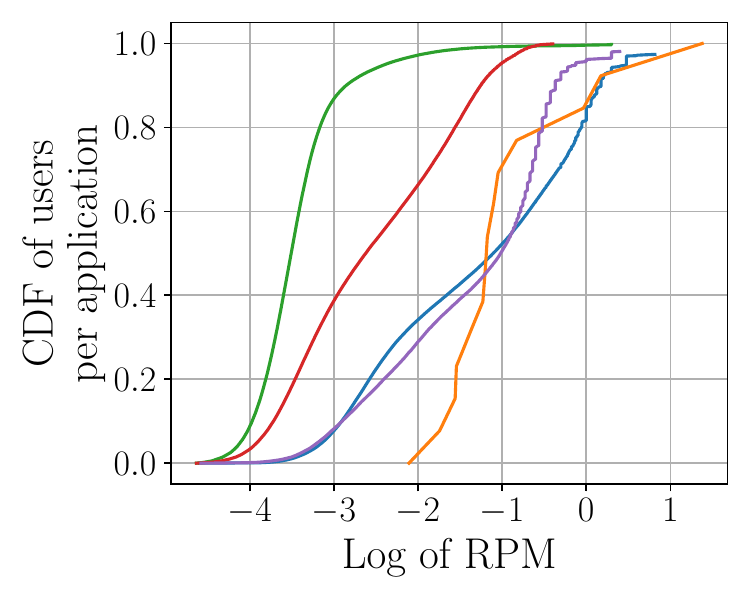}
        \caption{RPM across different users for different applications.}
        \label{fig:users_vs_rpm_app_new}
    \end{subfigure}
    \begin{subfigure}{0.24\textwidth}
        \includegraphics[width=\textwidth]{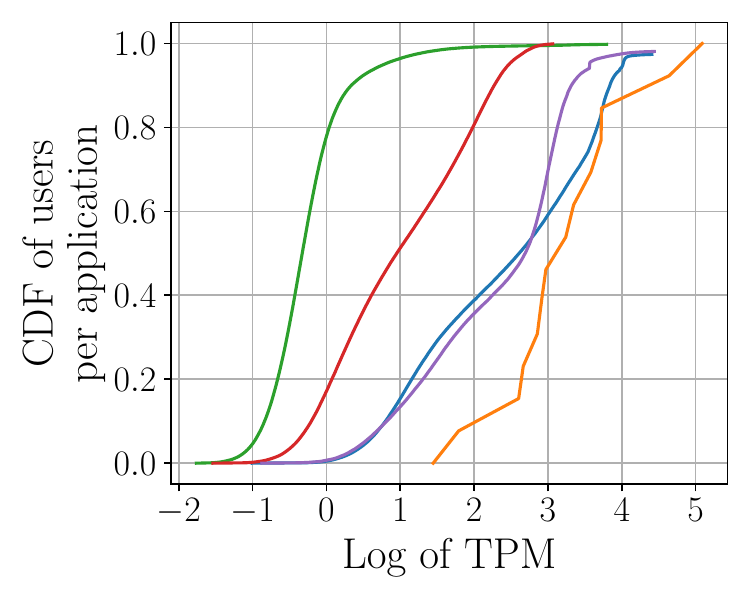}
        \caption{TPM across different users for different applications.}
        \label{fig:users_vs_tpm_app_new}
    \end{subfigure}
    \caption{Users' varying RPM and TPM across apps make prior policies ineffective for ensuring fairness and detering abusive behavior. 
    }
    \label{fig:user_behaviors}
\end{figure*}

We first investigate user request patterns, specifically assessing if requests from users and applications follow any particular trends with the goal to curb abusive behaviour and ensure fairness across users. 

First, we profile the CDF of users by requests per minute (RPM) and tokens per minute (TPM) to identify potential abusive behaviors.
Figures~\ref{fig:users_vs_rpm_new} and \ref{fig:users_vs_tpm_new} show that the overall request frequency and number of tokens sent by a user tends to fall in a certain range ($99^{th}$ percentile for RPM is below $10^2$). However, the maximum RPM and TPM observed is upto $10^4$ and $10^6$ respectively, showing signs that different users have different trends. In this case, a rate-limiting policy may mark a lot of users as abusive where as in reality this could be due to difference in their trends.  

Next, fine-grained analysis into the users of different applications reveal that users behave differently across applications in terms of RPM and TPM. For example, Figure~\ref{fig:users_vs_rpm_app_new} (each curve is for an app and each point corresponds to a user) shows that although one app has only around 20\% of its users' RPM to be less than or equal to $10^{-1}$, for others it could be well above 80\%. Again, Figure~\ref{fig:users_vs_tpm_app_new} shows that although around 80\% of the users of some app have a TPM of $10^2$ or less, for other apps it could be just 20\%. Both of these observations mark a stark dissimilarity in the RPM and TPM limits across applications, prompting us to treat users belonging to different applications differently. Hence, trying to maintain equal TPM across all users like VTC~\cite{vtc} in ensuring fairness might not be ideal and will not be sufficiently effective to deter abusive behavior.

\begin{thm}
    A user's overall RPM and RPM for each application falls in a certain range and these are different from one another. As benign users have different tokens per minute (TPMs), simply targeting equal TPMs across users is not sufficiently effective to deter the abusive behavior or achieve fairness.
\end{thm}

\subsection{Impact of Application Behavior}
\label{subsec:impact_application}

\begin{figure}
    \centering
    \begin{subfigure}{0.23\textwidth}
        \includegraphics[width=\textwidth]{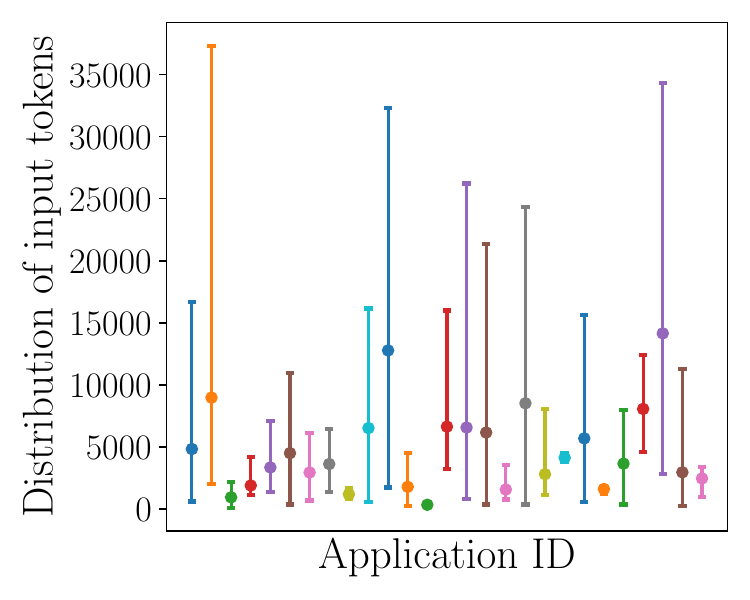}
        \caption{Input lengths across applications.}
        \label{fig:input_distribution_app}
    \end{subfigure}
    \begin{subfigure}{0.23\textwidth}
        \includegraphics[width=\textwidth]{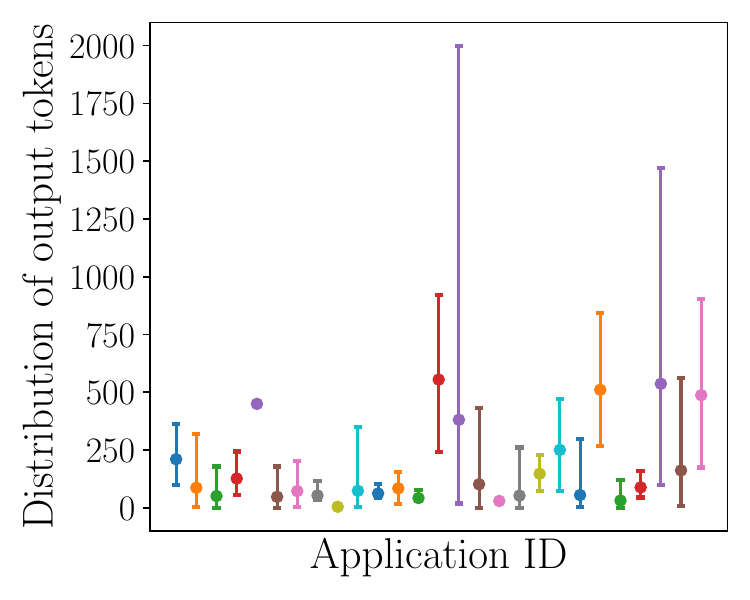}
        \caption{Output lengths across applications.}
        \label{fig:output_distribution_app}
    \end{subfigure}
    \caption{Token counts differ across applications, suggesting that LLM scheduling must consider variations in user applications.}
\end{figure}

\begin{figure}[t!]
    \centering
    \begin{subfigure}{0.23\textwidth}
        \includegraphics[width=\textwidth]{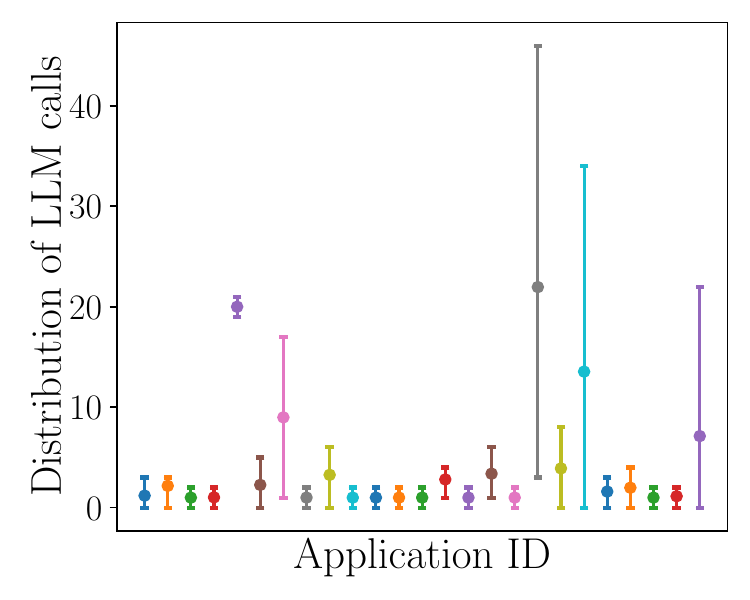}
        \caption{Number of calls across applications.}
        \label{fig:scenario_vs_llm_calls}
    \end{subfigure}
    \begin{subfigure}{0.23\textwidth}
        \includegraphics[width=\textwidth]{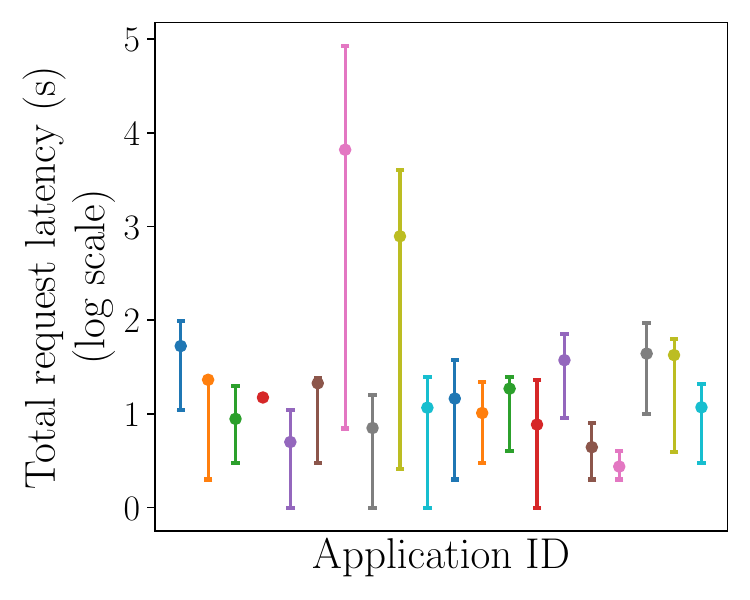}
        \caption{Latency in LLM calls across applications.}
        \label{fig:apps_vs_total_latency}
    \end{subfigure}
    \caption{The variability of LLM calls across apps must be considered to reduce latencies and queueing delays in multi-agent apps.}
\end{figure}

In this section, we closely examine the input, output, and total token lengths across various applications. Our aim is to detect token patterns that will aid in designing a fair weighting system for users of different apps. Figure~\ref{fig:input_distribution_app} illustrates that the average input lengths differ significantly across applications, with each having a specific range within which all its requests fall. Although Figure~\ref{fig:output_distribution_app} shows that the output lengths across applications are more similar than the input lengths, there are still some variations. 
This insight pushes us to reevaluate how we assess user service across different applications. We propose shifting from a uniform token weighting to a strategy that assigns distinct weights based on  application behaviors. For instance, consider two applications, A and B, where the average total request lengths are 10 and 50 tokens, respectively. If 5 tokens are processed for each application at a given moment, we calculate that users from app A and B have received 5/10 (i.e., 50\%) and 5/50 (i.e., 10\%) of their respective service as that is the portion of the request that gets completed for these users. This method differs from the prior fairness policy~\cite{vtc}, which would have allowed both users to receive an equal amount of service at that stage, and hence would treat them equally when scheduling requests.

\begin{thm}
Each application has its own normal range of input and output token lengths for requests, which vary across different applications.
This insight indicates that service calculations during LLM request scheduling must account for the application-level differences between user requests.  
\end{thm}

\subsection{Impact of LLM Agents and System Prompts}
\label{subsec:impact_llmagents}

\begin{figure}[t!]
    \centering
    \begin{subfigure}{0.23\textwidth}
        \includegraphics[width=\textwidth]{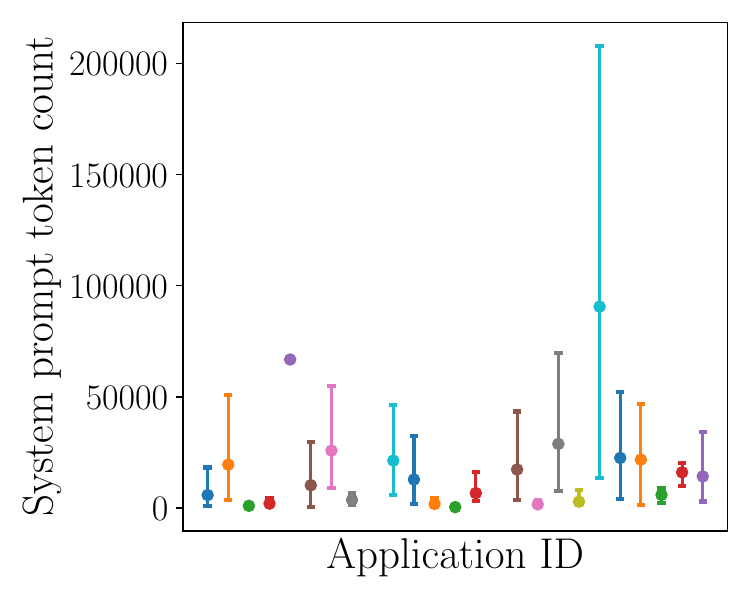}
        \caption{System prompt token in LLM calls across apps.}
        \label{fig:apps_vs_systemtokens}
    \end{subfigure}
    \begin{subfigure}{0.23\textwidth}
        \includegraphics[width=\textwidth]{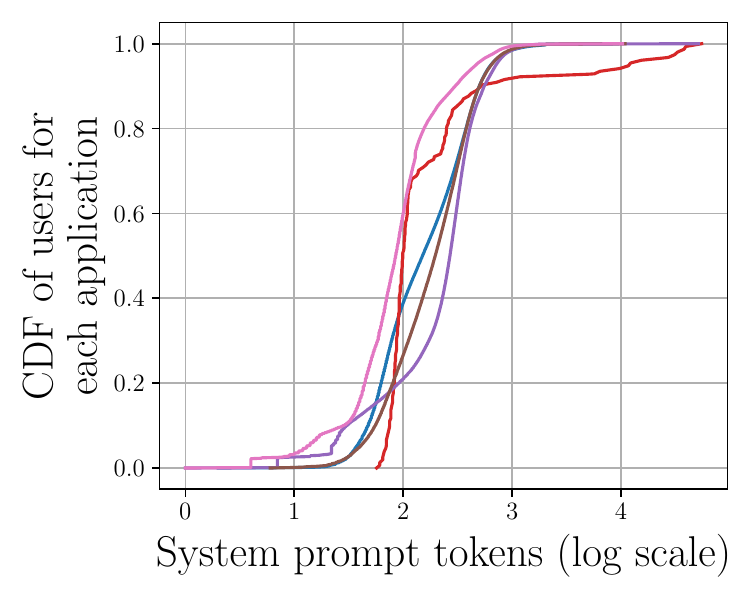}
        \caption{Output tokens in system prompts across users.}
        \label{fig:scenario_vs_output_tokens}
    \end{subfigure}    
    \caption{System prompts lead to varying characteristics in total tokens of each interaction and the number of output tokens}
\end{figure}

Next, we investigate the impact of LLM agents and system prompts on generating the final response for a user request. 
Figure~\ref{fig:scenario_vs_llm_calls} shows that across applications the number of LLM calls vary, i.e., the same application can generate different number of LLM calls (can be over 20) using LLM agents, 
%to generate a response for the user 
depending on the context of the user request. 
Furthermore, Figure~\ref{fig:apps_vs_systemtokens} shows that: (1) across applications the system prompt token count varies, and (2) even within the same application, the same LLM agents can generate different number of system tokens. 
Moreover, Figure~\ref{fig:scenario_vs_output_tokens} shows that these system prompts coming from LLM agents can lead to the generation of different number of output tokens. In particular, we observe that 20\% of the users of four applications are generating over 500 output tokens. Additionally, Figure~\ref{fig:apps_vs_total_latency} shows that these LLM calls have varied latencies (can be over 6 seconds), depending on the position of the LLM call request in the waiting queue for the interaction. 
%Such latencies 
Long latencies are undesirable for applications with tight SLOs, stressing the need for proper placement of LLM call requests during scheduling in order to reduce latencies and queueing delays for multi-agent LLM requests. 

\begin{thm}
    The number of LLM calls and their latencies vary across different applications and within the same application. Inefficient scheduling can cause these calls to get queued up, leading to delays in the execution of these LLM calls and hence results in high latencies. This observation indicates that scheduling must account for LLM call patterns to avoid delaying users. 
\end{thm}

\begin{figure*}[t!]
    \centering
    \begin{subfigure}{0.3\textwidth}
        \includegraphics[width=\textwidth]{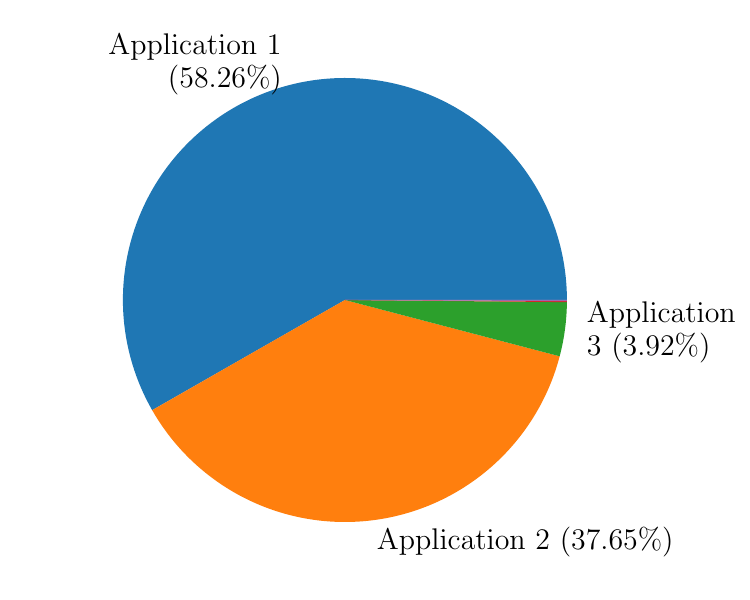}
        %\caption{Distribution of throttled requests between different applications.}
        \caption{Distribution of throttled requests across different applications.}
        \label{fig:piechart_throttled_users}
    \end{subfigure}
    \begin{subfigure}{0.3\textwidth}
        \includegraphics[width=\textwidth]{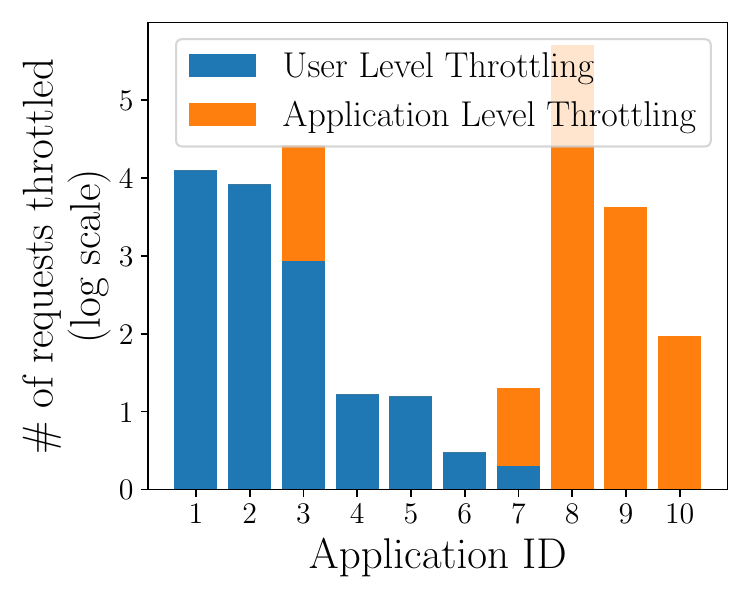}
        \caption{Ratio of requests throttled at application and user level for each application.}
        \label{fig:histogram_application_throttle}
    \end{subfigure}
    \begin{subfigure}{0.3\textwidth}
        \includegraphics[width=\textwidth]{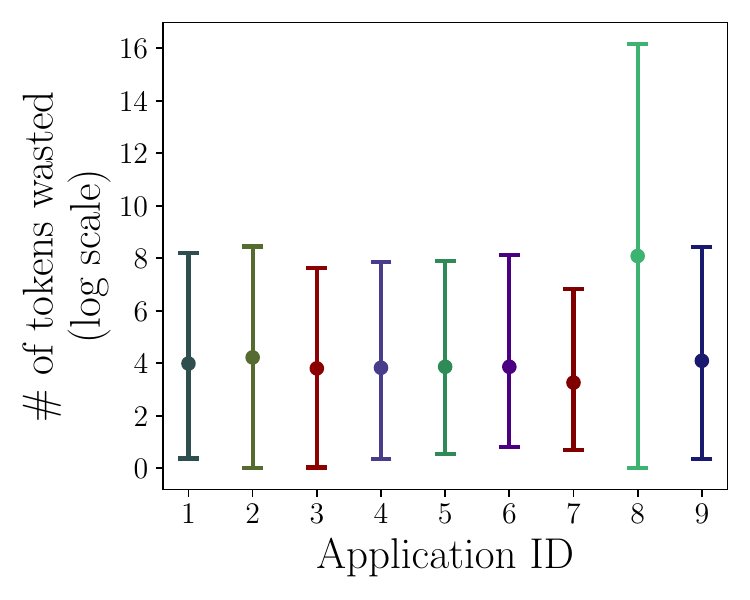}
        \caption{Distribution of wasted tokens due to throttling across applications.}
        \label{fig:scenario_vs_tokens_wasted}
    \end{subfigure}
    \caption{A throttling system without user and application-awareness can lead to resource wastage.} 
    \label{fig:rpm_throttling_effect}
\end{figure*}
\begin{table}[ht]
    \centering
    \caption{Distribution of graph sizes and tokens in graphs}
    \resizebox{0.9\columnwidth}{!}{
    \begin{tabular}{lrrrrr}
        \toprule
        Graph size & Requests & Avg Input & Avg Output & Avg Sum \\
        \midrule
        1 & 73.22\% & 4994.65 & 136.86 & 5131.52 \\
        2-10 & 26.09\% & 21352.13 & 241.19 & 21593.33 \\
        11-20 & 0.50\% & 51200.13 & 3258.75 & 54458.90 \\
        21-30 & 0.11\% & 65924.80 & 9217.13 & 75141.94 \\
        31-40 & $<$0.01\% & 94096.51 & 13582.78 & 107679.30 \\
        41-50 & $<$0.01\% & 161158.58 & 12874.49 & 174033.07 \\
        $>$50 & $<$0.01\% & 543559.74 & 22102.99 & 565662.74 \\
        \bottomrule
    \end{tabular}
    }
    
    \label{tab:graph_table}    
\end{table}

\subsection{Impact of RPM Throttling}
\label{subsec:impact_rpmthrottling}

In industries, the de-facto standard of throttling to curb abusive behavior has been to use rate-limiting solutions, i.e., if the number of requests belonging to a user goes above a certain limit, then the request gets throttled~\cite{claudepricing, gptpricing}. 
While simple, this basic throttling approach 
has several issues. First, as shown in Table~\ref{tab:graph_table}, around 26.8\% of all requests have a graph size larger than one (i.e., each interaction consists of multiple LLM calls), meaning that if the user goes over the limit within the interaction, the request will get throttled and the user will not get a response. Figure~\ref{fig:scenario_vs_tokens_wasted} shows that throttling unawareness across applications can waste over 30K tokens, diverting resources that could have otherwise reduced latency, improved throughput, and enhanced user satisfaction. 
Second, the needs of users may vary if they are receiving service from different applications. Hence, a fixed limit applied to all users might prevent certain users from receiving service if the number of LLM calls made by those users exceeds the predefined user-level request limit.  

Another variation of rate-limiting solution is to place limits on different applications~\cite{msappthrottling}, i.e., if the requests belonging to an application goes over a limit, all further user requests belonging to that application will get throttled. However, such a rate limiting approach will also have unintended consequences. First, a request limit for a specific application can overflow, causing unnecessary throttling of user requests despite unfilled quotas or low system load, resulting in inefficient resource utilization and user dissatisfaction. For example, Figure~\ref{fig:piechart_throttled_users} shows that three apps account for around 99.93\% of all  throttled requests, implying that their request limits are exceeded far more frequently than those of the other apps. 
Second, without enforcing user-specific limits, setting application-only limits allows malicious users to gain unfair access to the compute resources by flooding the system with requests across multiple applications. 

The nature of throttling can also vary across different applications. For example, Figure~\ref{fig:histogram_application_throttle} reveals that a large number of requests are frequently throttled at the application level (8, 9, and 10), at the user level (1, 2, 4, 5, and 6), and  a combination of both (3 and 7) although not all users from this throttled set was abusive. This result indicates that rate-limiting solutions are vulnerable to abuse. 

\begin{thm}
With the RPM policy, a certain number of users and applications tend to be throttled. Throttling in the middle of execution leads to the wastage of resources and users’ resource quota. 
This observation suggests the necessity of a user- and application-aware throttling policy to prevent abuse and minimize resource waste by throttling only during overloads and outside active LLM interactions.
\end{thm}

%% file: 4_FairServe.tex
\section{FairServe Design}
\label{sec:fairserve_design}

Our study in $\S$~\ref{sec:data_analysis} sheds light on the potential to address the problem of throttling in the middle of interactions and token wastage, and motivates the design of an application and user request characteristic-aware system for ensuring fairness. The main argument in VTC is to serve clients as per the principle of fair queueing~\cite{nagle1987packet}, where each client is guaranteed at least an equal amount of  resources, i.e., $1/n$ of the server's resources. 
%\sys argues in favor of weighted fair queueing~\cite{demers1989analysis}, where clients are guaranteed unequal amounts of server resources based on pre-determined weights. 
\sys argues for weighted fair queueing~\cite{demers1989analysis}, where resources are allocated based on pre-determined weights, allowing for an {\it equitable} allocation tailored to the specific needs or behavior of each user or application. {\it This approach recognizes that fairness is not always achieved through strict equality, as proposed by VTC~\cite{vtc}; instead, equitable allocation can better address heterogeneous demands of users and applications.}

Designing a weighted fair LLM serving system presents non-trivial challenges.  

\textbf{First}, in multi-tenant LLM service providers, resource wastage primarily arises from two areas: (1) allowing abusive users a portion of valuable resources, 
and (2) mid-interaction throttling even if system is underutilized. 
%Trying to increase resource utilization while maintaining quality service for benign users  and detecting abuse thus becomes challenging as that would require fine-grained tracking within an LLM interaction. 
Increasing resource utilization while ensuring quality service for benign users and detecting abuse thus becomes challenging, requiring fine-grained tracking within LLM interactions.
%and quantifying abusive behavior.

\textbf{Second}, users in multi-tenant environments belong to diverse applications, each with unique characteristics. Weighted fair allocation requires understanding each application's behaviors at different stages of an LLM interaction. 

This section presents the design principles of \sys to address these challenges, followed by the design detail. 

\begin{figure}[t]
\centering
\includegraphics[width=0.48\textwidth]{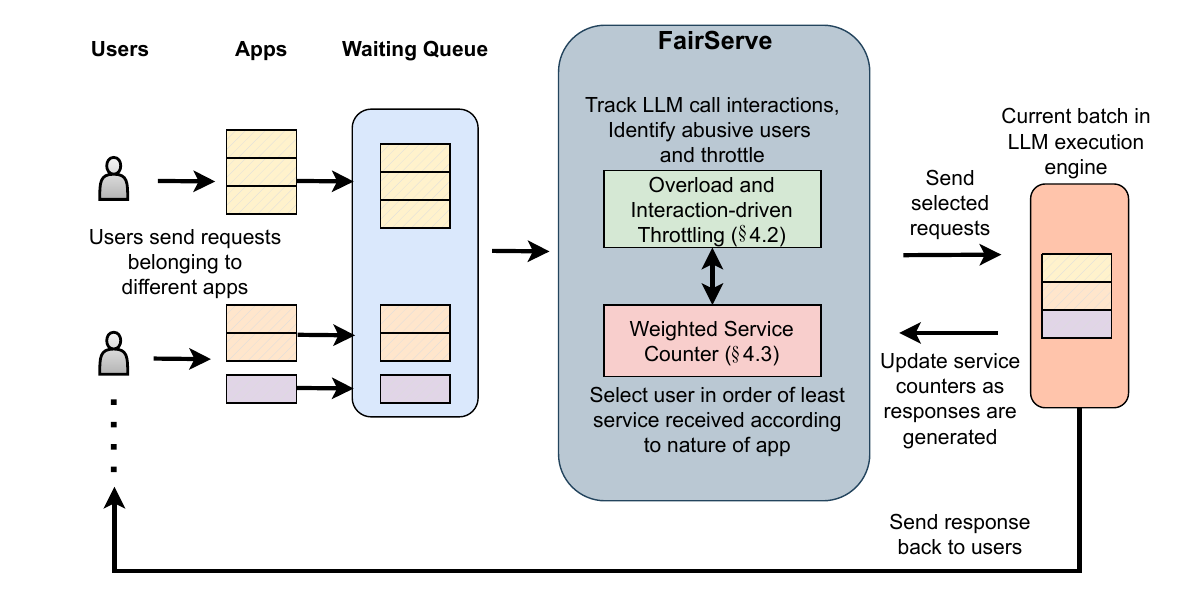}
\caption{The design overview of {\sys}. %\textcolor{red}{TODO: This fig needs to be referenced in Design text.}
}
\label{fig:FairServe_system_design}
\end{figure}

\begin{algorithm}[t!]
\caption{{\sys} System}
\label{secondComponent}
\begin{algorithmic}[1]
\STATE $B$: Dynamically changing current batch 
\STATE Service counter, $u_i \leftarrow 0$ for all user $i$
\STATE $Q$ $\leftarrow$ Dynamically changing waiting queue
\STATE $M_{new}$: New minibatch to be added to $B$
\STATE $T^r_g$: Global request throttling limit for all users
\STATE $T^r_a$: Request throttling limit for app $a$
\STATE $c^r_{g,u}$:  Count for requests for user $u$ 
\STATE $c^r_{a}$: Count for requests for app $a$
\STATE $\triangleright$ \texttt{Monitoring Stream:}
\WHILE{True}
    \IF {new request $r$ from user $u$ and app $a$ arrived}
        \IF {not $\exists r' \in Q, user(r')=u$}
            \IF {$Q=\emptyset$}
                \STATE let $e\leftarrow$ the most recent user to exit $Q$
                \STATE $u_u \leftarrow \max\{u_u, u_{e}\}$
            \ELSE
                \STATE $X \leftarrow \{i \mid \exists r' \in Q, user(r')=i\}$
                \STATE $u_u \leftarrow \max\{u_u, \min\{u_i \mid i\in X\}\}$
            \ENDIF
        \ENDIF
         %\State u $\leftarrow$ client(r)
    %\State a $\leftarrow$ app(r)
    \STATE $c^r_{g,u}$ $\leftarrow$ $c^r_{g,u}$ + 1 ; $c^r_{a,u}$ $\leftarrow$ $c^r_{a,u}$ + 1
    %\State $c^r_{a,u}$ $\leftarrow$ $c^r_{a,u}$ + 1
    \IF{system overloaded \& r not in interaction}
        \IF{ $c^r_{g,u}>T^r_g$}
            \STATE block $r$; break
        \ELSIF{$c^r_{a} > T^r_a$}
            %\If{$c^r_{a} > T^r_a$}
                \STATE block $r$; break
            %\EndIf
        \ENDIF
    \ENDIF
        \STATE $Q \leftarrow Q + r$
    \ENDIF
\ENDWHILE
\STATE $\triangleright$ \texttt{Execution Stream:}
\WHILE{True}
    \IF{can\_add\_new\_request()}
        \STATE $M_{new} \leftarrow \emptyset$
        \WHILE{True} 
            \STATE $Y = find\_incomplete\_interactions(B)$
            \IF{any interaction is incomplete in B}
                \STATE $k \leftarrow \argmin_{i \in \{user(r) \mid r\in Y\}} u_i$
                \STATE x $\leftarrow$ incomplete interaction from k in B 
                \STATE $r$ $\leftarrow$ next request in $x$.
            \ELSE
                \STATE let $k \leftarrow \argmin_{i \in \{user(r) \mid r\in Q\}} u_i$
                \STATE let $r$ be the earliest request in $Q$ from $k$.
            \ENDIF
            \STATE $M_{new} \leftarrow M_{new} + r$; $Q \leftarrow Q - r$
            %\State $Q \leftarrow Q - r$
        \ENDWHILE
        \STATE forward\_prefill($M_{new}$)
        \STATE $B \leftarrow B + M_{new}$
    \ENDIF
    \STATE forward\_decode($B$)
    \STATE $F \leftarrow$ find\_finished\_requests($B$)
    \FOR{each f in F}
        \STATE $i$ $\leftarrow$ user(f); $a$ $\leftarrow$ application of request $f$
        \STATE $j$ $\leftarrow$ current stage of interaction of f
        %\STATE $a$ $\leftarrow$ application of request $f$
        \STATE $w_{aj}$ $\leftarrow$ weight of m in stage j
        \STATE $u_i \leftarrow u_i + E_i( \frac{\alpha L^{ai}_{I} + \beta L^{ai}_{S} + \gamma L^{ai}_{O}}{w_{aj}})$
    \ENDFOR
\ENDWHILE
\end{algorithmic}
\label{alg:wsc}
\end{algorithm}

\subsection{\sys Overview}
\label{subsec:overview}
\sys comprises of two core components: (1) Overload and Interaction-driven throttling (OIT), and (2) Weighted Service Counter (WSC) scheduling (see Figure~\ref{fig:FairServe_system_design}).  
In contrast to traditional RPM-based throttling, OIT throttles user requests only when the KV cache is overloaded, thus making maximum utilization of available resources. 
{\sys} uses a combination of user- and application-level limits to perform throttling at the LLM interaction level instead of at the request level, reducing token wastage by accounting for user and application behaviors. 
WSC crafts a user service weighing mechanism determined by the ratio of tokens processed to the expected token count (e.g., the average based on historical statistics) for the application associated with each user's requests at each level of the LLM interaction. WSC selects requests from the users who have received the least service, defined by a weighted resource slice. This weight is calculated based on the token ratio to ensure fairness across users.

Algorithm~\ref{alg:wsc} shows the details regarding how the two components interact with one another while processing requests. The entire system is integrated with the state-of-the-art continuous batching mechanism and operates in two parallel streams--- the monitoring and the execution stream. The monitoring stream continuously listens for incoming requests and the execution stream is in charge of the prefill and the decode phases. OIT is a part of the monitoring stream (Alg.~\ref{alg:wsc}, lines 19-25) and the larger chunk of weight and service allocation mechanisms in WSC take place in the execution stream (Alg.~\ref{alg:wsc}, lines 27-48).

\subsection{Overload \& Interaction-driven Throttling (OIT)}
\label{subsec:odt}

OIT performs two main functions: (1) tracking requests from users and applications against specified limits at both the user  and application level; and (2) tracking the condition of the KV cache and throttling based on the specified limit for users and apps only when the KV cache is overloaded. 

To track user and application requests, OIT maintains a dictionary that maps each user and application to the arrival times of their respective requests. OIT takes a combined rate-limiting approach that merges both user and application limits, based on the analysis of historical data. A combined user and application-aware approach helps address the challenge of resource wastage caused by abusive behaviors. When a new request arrives, its arrival time is appended to the list associated with the corresponding user and app. Tracking arrival times enables OIT to detect if any user or application exceeds their request limit. Upon request arrival, the internal request counter for the corresponding user and application is incremented (Alg.~\ref{alg:wsc}, line 19). 

OIT continuously monitors the state of the KV cache and assesses whether new requests can be added to the waiting queue.
When the KV cache becomes overloaded, OIT checks whether a request is in the middle of an interaction. If so, this request is not throttled and instead stalled to (1) guarantee resource utilization and (2) minimize unnecessary token wastage. 
Otherwise, at first, the user's RPM, and later the application's RPM limit, is checked to see if the user or app has exceeded the limit before the incoming request is served. If either limit is exceeded, the next request is throttled (Alg.~\ref{alg:wsc}, lines 21-24). 
The requests in the waiting queue are added to a new, pending batch of requests and then merged into the active batch currently being served (Alg.~\ref{alg:wsc}, line 25).

\subsection{Weighted Service Counter (WSC)}
\label{subsec:wsc}
WSC performs two main functions: (1) calculating the service received by users based on the ratios of input, system, and output tokens processed for their requests to the maximum token limits of the corresponding apps; and (2) identifying the user with the least service received and forwarding the user's request for processing.

To align user service with the characteristics of their corresponding applications, WSC updates the service received by users based on the ratio of processed tokens to the normal range (average) of request token length associated with the app. Before a user receives a response, the request goes through multiple stages of an interaction depending on the application. We model the weight of app 'a' in stage 'j' of interaction in Eq.~\ref{eq:appweight}. 

\begin{equation}
%\begin{split}
 w_{aj} = \alpha\bar{N}^{aj}_I+ \beta \bar{N}^{aj}_S + \gamma \bar{N}^{aj}_O  
 % \\
 % \hspace{-5cm}  w_i = \sum_{i=1}^{m_i}c_{ij}
%\end{split}
\label{eq:appweight}
\end{equation}

Each of the different types of tokens also has a weight associated with it. Since processing input tokens can be parallelized and hence is much fast compared to processing the output tokens, weights enable WSC to ensure that service is counted according to the number of different categories of tokens that are getting processed. Following OpenAI conventions, we fix $\alpha = 1$, $\beta = 2$, and $\gamma = 1$ to be the weights for input, system and output tokens respectively. $\bar{N}^{aj}_I$, $\bar{N}^{aj}_S$, and $\bar{N}^{aj}_O$ represent the expectation of the number of input, system, and output tokens respectively for the app a at stage j of an interaction. We calculate these expectation values based on analysis on historical data.

The input, system prompt, and output lengths for each user $i$ corresponding to app $a$ are denoted by $L^{ai}_{I}$, $L^{ai}_{O}$, and $L^{ai}_{S}$ respectively.
The total service received by a user at the end of an interaction is computed using Eq.~\ref{eq:userservice} where app $a$ requires $m_i$ LLM calls to complete an interaction and $E$ is a user priority factor that can be adjusted to give preference to certain users based on system policies or usage trends.

\begin{equation}
S_i^{a} = E_i * ( \sum_{j=1}^{m_i} \frac{\alpha L^{aij}_{I} + \beta L^{aij}_{S} + \gamma L^{aij}_{O}}{w_{aj}})   
\label{eq:userservice}
\end{equation}
 
WSC maintains a service counter $u_i$ for all users which is initialized to 0. 
Waiting queue, Q immediately adds a new incoming request upon its arrival after checking for abusive behaviour using overload-driven throttling $\S$~\ref{subsec:odt} (Alg.~\ref{alg:wsc}, lines 19-24). Q is a dict that maps users to their corresponding incoming requests. If this is the only request from the sender user, then a counter adjustment takes place (Alg.~\ref{alg:wsc}, lines 12-18) similar to prior research~\cite{vtc}. This sort of adjustment is done to create balance, ensuring that underloaded periods of certain users do not create unfairness against other active users. 

In the execution stream, WSC evaluates whether new minibatches, $M_{new}$, composed of user requests, can be merged with the current batch, $B$. Initially, it identifies which requests within the current batch are part of an ongoing interaction, requiring additional processing to complete the interaction. Among those requests, the system identifies the user who has received the least service so far and selects the next request from that user's interaction  (Alg.~\ref{alg:wsc}, lines 31-35). This technique enables \sys to keep lower queueing lengths and ensure better response times. 

Otherwise if no requests in the current batch are part of an ongoing interaction, the earliest request from the user who has received the least service, as determined by Eq.~\ref{eq:userservice}, is selected for inclusion in $M_{new}$ (Alg.~\ref{alg:wsc}, lines 36-39). Once the prefill and decode stages for a request are completed, the service counters for the users are updated, proportional to the app's assigned weights (Alg.~\ref{alg:wsc}, lines 40-48).

%% file: 5_Evaluation.tex
\section{Evaluation}

\begin{figure*}[t!]
    \centering
    \begin{subfigure}{0.24\textwidth}
\includegraphics[width=\textwidth]{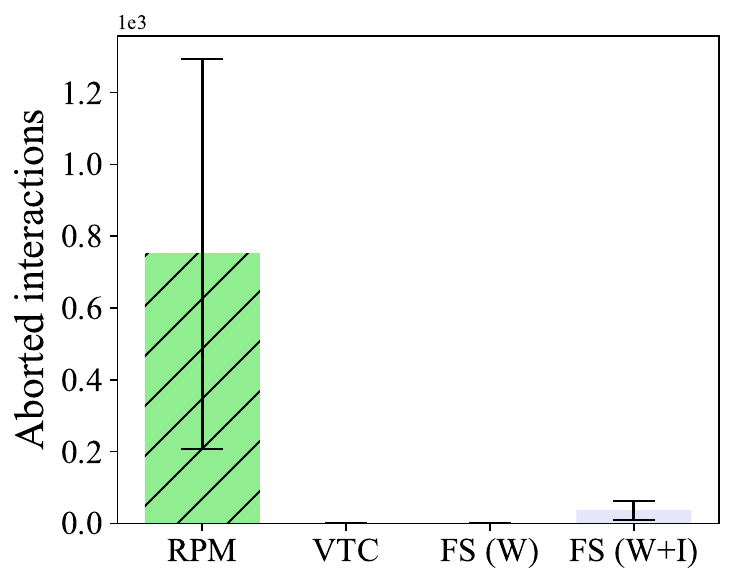}
        \caption{Aborted interactions during LLM serving.}
        \label{fig:aborted_interaction_rpmvsfs}
    \end{subfigure}
    \begin{subfigure}{0.24\textwidth}
        \includegraphics[width=\textwidth]{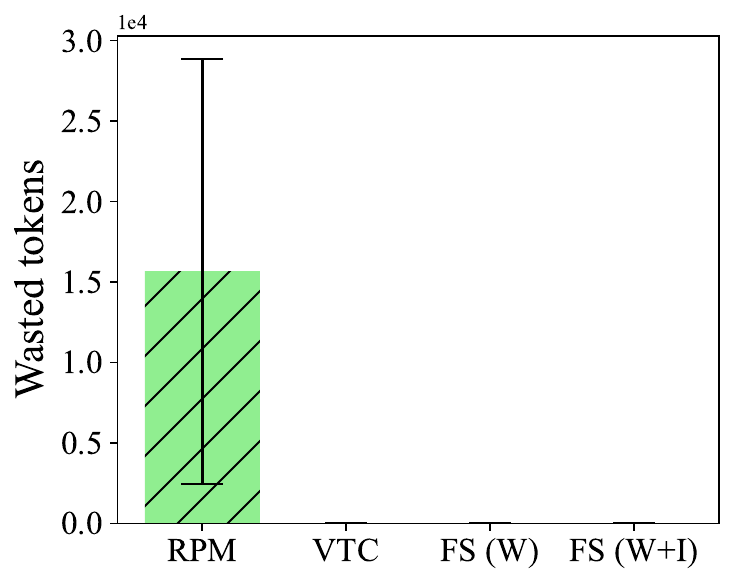}
        \caption{Tokens wasted from interaction throttling.}
        \label{fig:wasted_token_rpmvsfs}
    \end{subfigure}
    \begin{subfigure}{0.24\textwidth}
        \includegraphics[width=\textwidth]{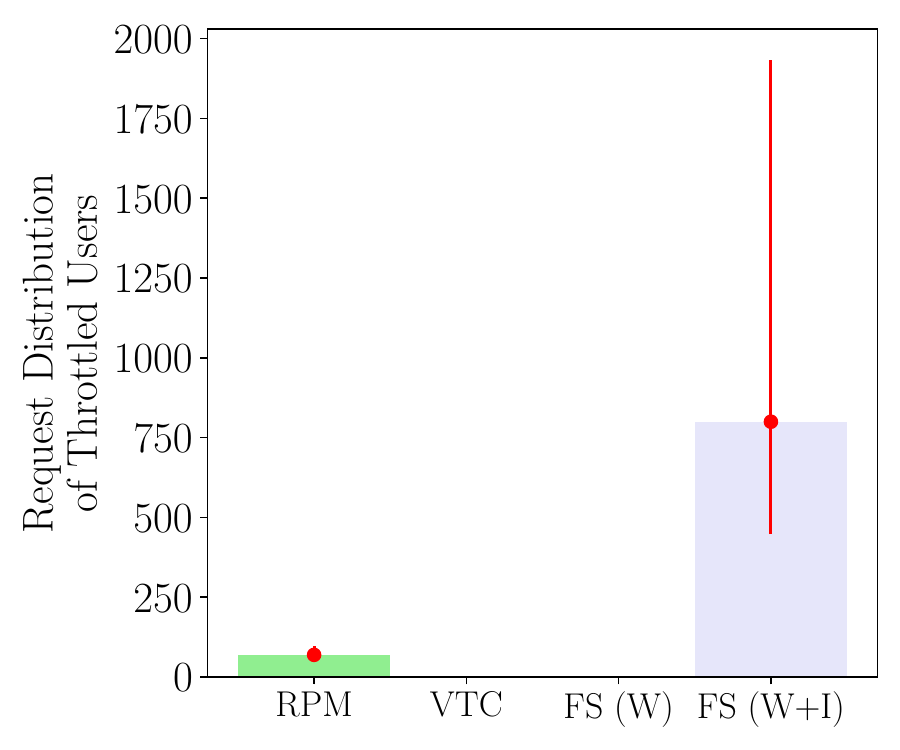}
        \caption{Request distribution within throttled users.}
        \label{fig:throttled_user_request_distribution}
    \end{subfigure}
    \begin{subfigure}{0.24\textwidth}
        \includegraphics[width=\textwidth]{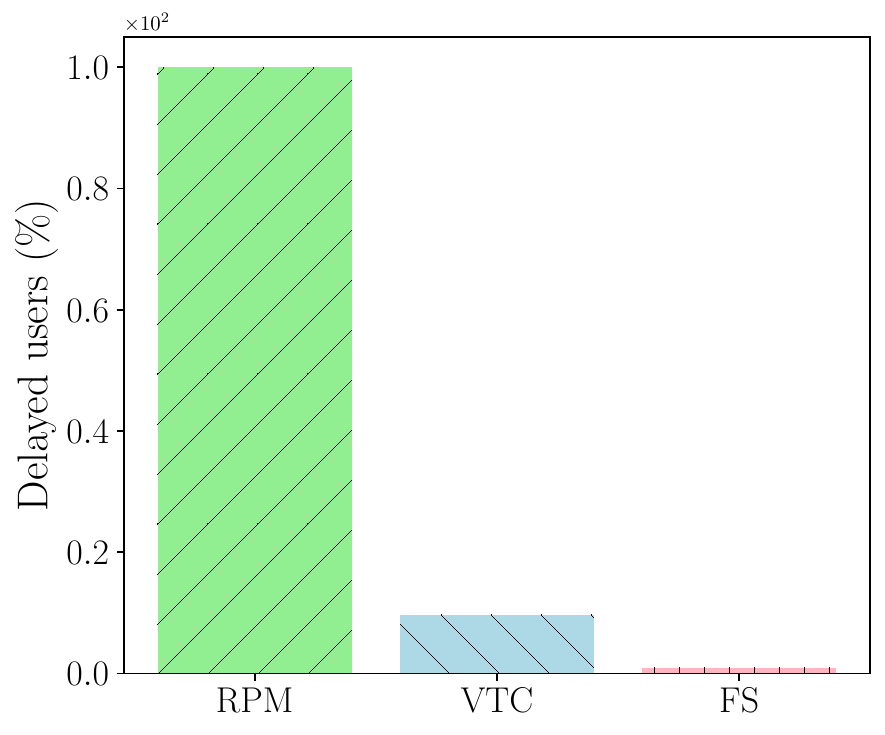}
        \caption{Delayed users due to longer waiting queue lengths.}
        \label{fig:delayed_users}
    \end{subfigure}
    \caption{\sys does not incurr resource wastage as is it is multi-agent LLM interaction-aware.}
    \label{fig:abortedandwasted}
\end{figure*}

In this section we evaluate the effectiveness of \sys against other baselines on the real-world workload trace that we analyzed in $\S$~\ref{sec:data_analysis}. We run all of the policies for different durations of time and observe how each perform on curbing abusive behaviour and achieving fairness. FS (W+I) or FS indicates \sys with all components and FS (W) indicates \sys is using only the WSC component. We denote all of the applications in our experimentation with Ids to protect anonymity and privacy.

Our baselines are two state-of-the-art solutions used widely in industry and academia for curbing abusive behavior and achieving fairness---(1) RPM (Requests per minute), i.e., a rate-limiting solution; and (2) VTC~\cite{vtc}.
Our evaluation aims to answer the following questions:

\begin{itemize}
    \item To what extent can FS curb abuse? ($\S$~\ref{eval:curbabusive})
    \item How effectively does FS reduce queuing delays for users in multi-agent LLM apps? ($\S$~\ref{eval:queuedelays})
    \item What is the impact of FS on throughput and time-to-first-token (TTFT) latency? ($\S$~\ref{eval:throughputandlatency})
    \item Can FS improve the user serving experience? ($\S$~\ref{eval:servedexperience}) 
\end{itemize}

\subsection{Curbing Abusive Behaviour}
\label{eval:curbabusive}
As discussed in $\S$~\ref{sec:background}, in multi-agent LLM apps, user queries initiate interactions having multiple LLM calls. Throttling in the middle of interaction wastes resources on tokens that provide no benefit, as the user receives no final response. Figure~\ref{fig:aborted_interaction_rpmvsfs} shows that as RPM based strategy is completely oblivious of interactions, it throttles users blindly based on a specified limit, and hence throttles $21.15\times$ more interactions compared to interaction-aware FS (W+I). FS (W+I) throttles only those interactions that are not in the middle of execution. Figure~\ref{fig:wasted_token_rpmvsfs} shows that RPM's unwareness regarding multi-agent apps wastes a lot of tokens (mean: $15.66*10^{3}$). On the other hand, FS (W+I), with multi-agent awareness and knowledge about the KV cache, only throttles during system overloads when users are not in mid-interaction, thus not incurring any token wastage while maintaining maximum KV cache utilization. 

Moreover, FS (W+I) manages to throttle only users who fall outside the normal range, i.e., abusive. Figure~\ref{fig:throttled_user_request_distribution} shows that while FS (W+I) throttles users whose request distribution is well above the normal range, RPM policy throttles users with very low request distribution. VTC and FS (W) does not have any throttling policy and hence serves all users without throttling. However, our evaluation shows that VTC wastes approximately $77.2 \times 10^3$ tokens by serving both abusers and benign users equally.

\begin{figure}[t!]
    \centering
    \begin{subfigure}{0.23\textwidth}
\includegraphics[width=\textwidth]{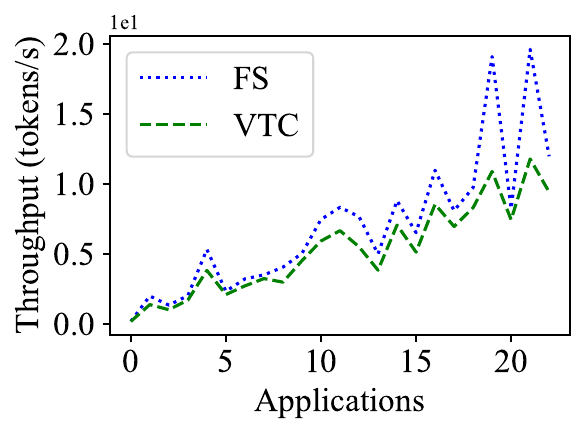}
        \caption{Throughput variation.}
        \label{fig:appvsthroughput}
    \end{subfigure}
    \begin{subfigure}{0.23\textwidth}
        \includegraphics[width=\textwidth]{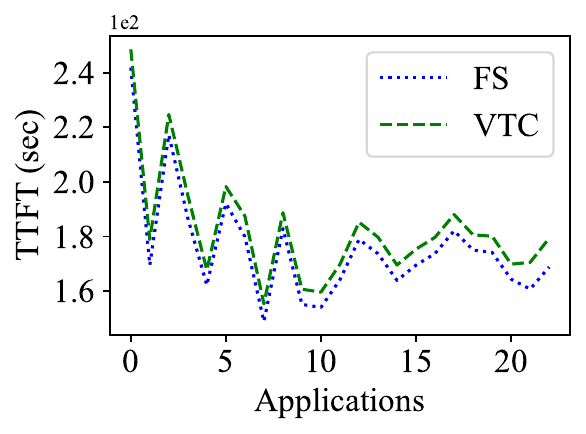}
        \caption{TTFT variation.}
        \label{fig:appvsttft}
    \end{subfigure}
    \caption{\sys (FS) maintains a higher throughput and a lower TTFT (time-to-first-token) latency across apps.}
    \label{fig:appandthroughputttft}
\end{figure}

\subsection{Reducing Queueing Delays}
\label{eval:queuedelays}
One of the aspects that adversely affects user experience is if requests are waiting for longer durations before processing. Since, \sys is multi-agent aware, it prioritizes serving users who are already in the middle of an interaction using the WSC component ($\S$~\ref{subsec:wsc}). Thus users do not wait a long time before receiving the final response. Figure~\ref{fig:delayed_users} shows that only a mere $0.93\%$ users experience queueing delays in \sys, making it $10.67\times$ and $93\times$ less compared to VTC and RPM respectively.  

\subsection{Improvement in Throughput and Latency}
\label{eval:throughputandlatency}
Since \sys manages to curb abusive behavior, we now look at how such curbing impacts the throughput of the user requests across the different applications. 

Figure~\ref{fig:appvsthroughput} shows that \sys maintains $1.03\times$-$1.75\times$ higher throughput across all of the applications. Moreover, Figure~\ref{fig:appvsttft} shows that \sys manages to attain $1.03\times$-$1.06\times$ lower TTFT latency, i.e., the time required to generate the first token in a request across all of the apps. \sys obtains this improvement obtained in throughput and TTFT latency as less users experience queueing delays due to WSC component and abusive users are not given opportunity to waste valuable resources and time due to the OIT component.

We further dig deeper into the throughput of user requests to understand the impact of \sys's policies at individual stages of the LLM inference process. Table~\ref{tab:TPM_req_served} shows that FS (W+I) maintains a $1.05\times$ and $1.1\times$ higher throughput compared to VTC and RPM respectively in the prompt processing phase. In the decode phase, FS (W+I) maintains a $1.02\times$ and $1.47\times$ higher throughput compared to VTC and RPM. Although TTFT is the same for \sys and VTC, we observe that it is  lesser for RPM. As RPM throttles lots of requests blindly, the requests that do get processed enjoys a slightly %shorter 
better TTFT and TBT. 

\subsection{Improvement in Served Experience.}
\label{eval:servedexperience}
In a real-time LLM serving system, requests are processed continuously. When incoming requests exceed the server’s capacity, an RPM policy may throttle users or requests but still handles a high volume due to the abundance of incoming requests. Here, we analyze the proportion of users receiving a final response among all users who received feedback. We define {\it served users} as those who received a final response and {\it service provided} as the requests or interactions that led to a final response without being throttled.

Table~\ref{tab:TPM_req_served} shows that \sys consistently has a higher percentage of served users and provided service compared to RPM. In particular, for multi-agent apps involving interactions, it manages to provide 100\% service and served user indicating that \sys could improve the user experience for all users in multi-agent scenarios. Note that, we do not show this comparison against VTC as it does not throttle, and hence all users receive feedback even if it is at the expense of increasing the latency of another user.

\subsection{A Case Study: Comparing Diverse Apps}

\autoref{tab:application_comparison} shows three diverse applications sending requests with a timeout. Application 14 has request with moderate prompt size and heavy decode size, application 7 has prompt heavy requests with moderate sized decodes while application 12 has both, light prompts and decodes. We see that a naive RPM strategy with FCFS scheduling disproportionately benefit applications 14 and 7 due to their heavy nature, as observed in their prompt and decode throughputs. FS (W) and VTC are able to balance the service each of these applications is getting, improving the prompt throughput for application 12 by 24$\times$. However, they still waste a lot of tokens while scheduling as they ignore multi-agent requests in these applications. Finally, we observed that while VTC is wasting 3224 tokens by ignoring multi-agent calls, FS (W+I) is able to provide service of comparable levels to each applications, while wasting almost $8\times$ (461) less tokens.% in the process.

\input{plots/evaluation/TPM_Req_Served_tablel}

\begin{table}
    \centering
    \caption{Case study: Comparison of service provided to specific applications by each strategy 
    % {\color{red}TODO: tokens wasted row looks ugly, make it better}
    }
    \resizebox{0.80\linewidth}{!}{%
        \begin{tabular}{lcccc}
        \toprule
            - & & Application 14 & Application 7 & Application 12 \\
        \midrule
            Requests & & 269 & 139 & 448 \\
            Avg. Prompt &  & 6370 & 14854 & 999  \\
            Avg. Decode & & 102 & 74 & 32 \\
            \midrule
            \multirow{4}*{Requests Served} & RPM & 56 & 4 & 2  \\
             & VTC & 37 & 29 & 73  \\
             & FS (W) & 37 & 29 & 73   \\
             & FS (W+I) & 37 & 27 & 80  \\
             \midrule
            \multirow{2}*{Prompt Throughput} & RPM & 360.99 & 859.70 & 5.19 \\
             & VTC & 256.14 & 374.31 & 124.19  \\
             & FS (W) & 253.41 & 390.47 & 124.69  \\
             & FS (W+I) & 258.20 & 359.02 & 127.58  \\
            \midrule
        \end{tabular}
    }
    \label{tab:application_comparison}
\end{table}

%% file: plots/evaluation/TPM_Req_Served_tablel.tex
\begin{table}[t]
    \centering
    \caption{Throughput and latency during real-time processing.}
    \resizebox{0.99\linewidth}{!}{
        \begin{tabular}{lccccccccc}
        \toprule
         \multirow{2}*{Scheduler} & \multicolumn{2}{c}{Throughput (TPS)} & \multirow{2}*{TTFT} & \multicolumn{2}{c}{Service provided \%} &  \multicolumn{2}{c}{Served users \%} \\ 
          & Prompt & Decode & & Requests & Interactions & Requests & Interactions \\
\midrule
RPM & {11155.15} & {182.85} & 40.20 & 51.52 & 85.36 & 92.03 & 87.64  \\
VTC & {12190.10} & {262.97} & 42.57 & - & - & - & - \\
FS (W)& {12217.11} & {267.05} & 42.82 & - & - & - & -  \\
FS (W+I) & {12247.98} & {267.99} & 42.53 & 96.67 & 100 & 99.45 & 100 \\
\bottomrule
        \end{tabular}
    }
    \label{tab:TPM_req_served}
\end{table}

%% file: 6_RelatedWork.tex
\section{Related Work}
\label{sec:related_work}

\noindent\textbf{Fairness \& Characteristic-awareness in ML Training.}
Fair scheduling mechanisms like Quincy~\cite{isard2009quincy}, DRF~\cite{DRF}, and CARBYNE~\cite{carbyne} have been proposed for long-running jobs in clusters. 
However, ML jobs differ from traditional workloads in their nature, with specific requirements for scheduling and placement. As a result traditional policies fail to ensure fairness~\cite{mahajan2020themis}. 

To ensure fairness in ML job scheduling, Gandiva~\cite{gandiva} introduces a profiling and resource-trading approach, Themis~\cite{mahajan2020themis} employs an auction-based method to balance fairness and efficiency, Gavel~\cite{gavel} combines optimization techniques with preemptive, round-based scheduling to fairly allocate resources among users, and Pollux~\cite{qiao2021pollux} dynamically adjusts resources to optimize both cluster-wide performance and fairness. Sia~\cite{sia} leverages integer linear programming to ensure fairness while enabling elastic scaling for hybrid parallel ML jobs. Additionally, SHADE~\cite{khan2023shade} exploits sample characteristics and FedCaSe~\cite{khan2024fedcase} exploits client and sample characteristics to speed up ML training. 

While these works address scheduling, fairness, and efficiency in ML training employing a wide variety of ML characteristic-aware techniques, they are not designed to handle the specific requirements of LLM serving jobs in multi-tenant environments. In such settings, LLM serving requires addressing distinct issues, such as prioritizing users and applications during request batching and throttling to mitigate abusive behavior.

\noindent\textbf{Fairness \& Characteristic-awareness in LLM Serving.} Recently, there has been growing momentum in both industry and academia focused on enhancing LLM serving systems. Techniques like batching~\cite{agrawal2023sarathi}, memory optimizations~\cite{s3},
%~\cite{s3,flashinfer}
scheduling~\cite{wu2023fastserve}, exploitation of other LLM-specific characteristics like chunking prefills~\cite{agrawal2024sarathiserve}, disaggregation of execution stages~\cite{splitwise,zhong2024distserve} or a combination of these techniques~\cite{vllm,yu2022orca,deepspeed-fastgen} have been proposed to improve the throughput and latency of serving models across users.

Although these systems aim to enhance LLM inference performance for users, the critical issue of ensuring fairness in serving LLMs has remained largely unexplored.  VTC~\cite{vtc} introduces a scheduling policy designed to ensure that users receive service in a fair and equal manner. However, in multi-tenant environments, where users request services from diverse applications, enforcing equal service across users is suboptimal.
Different applications have varying resource demands that must be met for optimal performance. \sys bridges this gap with an application-aware fair scheduling policy that balances fairness with each application's specific needs.

%% file: 7_Conclusion.tex
\section{Conclusion}
\label{sec:conclusion}
While there have been significant strides in lowering latency and enhancing throughput for LLMs, ensuring fairness across diverse LLM application users is a newer challenge that has garnered significant attention. While several fairness methods have been suggested, they fall short when applied to LLM users in diverse applications. In this work, we conduct a large-scale analysis of LLM user and application characteristics at a leading LLM service provider. Using our gathered insights, we propose \sys---a user and LLM application characteristic-aware LLM serving system that ensures fairness in scheduling user requests in a multi-tenant setting.